\documentclass{article}

\usepackage{arxiv}

\usepackage[utf8]{inputenc} 
\usepackage[T1]{fontenc}    
\usepackage{natbib}
\usepackage{hyperref}       
\usepackage{url}       
\usepackage{orcidlink}
\usepackage{booktabs}       
\usepackage{amsfonts}       
\usepackage{nicefrac}       
\usepackage{microtype}      
\usepackage{xcolor}         
\usepackage{graphicx}       
\usepackage{amsmath, amssymb}        
\usepackage[ruled,vlined]{algorithm2e}
\usepackage{subcaption}
\graphicspath{ {./Images/} }

\newcommand{\ket}[1]{\left| #1 \right\rangle}

\title{Supervised Latent Restructuring for Small-Data Quantum Learning in Plant Phenomics}

\author{%
  Alakananda Mitra\, \orcidlink{0000-0002-8796-4819} \\
  Nebraska Water Center, IANR\\
  University of Nebraska--Lincoln\\
  Lincoln, NE 68588 \\
  \texttt{amitra6@unl.edu} \\
  \And
  David H. Fleisher\, \orcidlink{0000-0002-0631-3986}, Vangimalla Reddy \\
  Adaptive Cropping Systems Laboratory \\
  USDA-ARS \\
  Beltsville, MD 20705 \\
  \texttt{\{david.fleisher, vangimalla.reddy\}@usda.gov} \\
  \And
  Chittaranjan Ray\, \orcidlink{0000-0002-1731-2030} \\
  Nebraska Water Center, DWFI \\
  University of Nebraska--Lincoln \\
  Lincoln, NE 68588 \\
  \texttt{cray@nebraska.edu} \\
}

\begin{document}

\maketitle

\begin{abstract}
High-dimensional biological data often exhibit a severe mismatch between feature dimensionality and sample size, making reliable classification difficult in extremely small-data regimes. In these settings, kernel methods can lose discriminative power when latent compression fails to preserve class-separating structure. We study this problem in fine-grained plant phenomics and propose a hybrid workflow that compresses 1280-dimensional deep image embeddings into a 64-dimensional PCA space and then restructures them into an 11-dimensional supervised latent space using Linear Discriminant Analysis (LDA), followed by GPU-accelerated Quantum Kernel Alignment (QKA) on NVIDIA L40S hardware. Empirically, supervised latent restructuring substantially improves the geometric separability of the compressed representation, increasing the Silhouette coefficient from 0.003 in the raw embedding space and -0.006 in PCA-64 to 0.197 in the supervised LDA-11 space. However, downstream classical evaluation reveals a clear compression trade-off: Linear SVM and XGBoost improve in the restructured latent space, whereas RBF-SVM and Random Forest degrade under the same 11-dimensional bottleneck. Under a constrained optimization budget, QKA in this regime remains challenging, indicating that latent geometry alone is not sufficient for strong trainable quantum performance. These findings position representation geometry as a central design variable in small-data quantum learning and expose the practical difficulty of recovering nonlinear discriminative structure from aggressively compressed biological representations.
\end{abstract}

\section{Introduction}
\label{Sec:intro}
High-dimensional biological sensing data often present a severe mismatch between feature dimensionality and sample size. This challenge is especially pronounced in plant phenomics, where disease recognition must often be performed from limited labeled data despite substantial visual complexity across classes. In fine-grained plant pathology, diagnostically meaningful differences may appear through subtle lesion texture, discoloration, or morphology rather than large global changes, making reliable classification difficult even for strong machine learning models \cite{mohanty2016using, ferentinos2018deep, barbedo2018impact}. Under such conditions, the challenge is not only learning from a few examples but also preserving biologically meaningful structure when high-dimensional representations are compressed for downstream inference.

This issue becomes particularly important in quantum machine learning (QML). Many practical quantum pipelines are fundamentally qubit-limited: they cannot directly process high-dimensional classical embeddings and therefore require aggressive dimensionality reduction before quantum encoding \cite{havlicek2019supervised, schuld2019quantum}. In practice, this creates a difficult trade-off. Compression is necessary to make quantum learning feasible, but it may also remove class-separating information that is essential for reliable classification. As a result, the success of quantum kernel methods may depend not only on the quantum circuit or optimizer, but also on how the classical representation is restructured before it is mapped into Hilbert space \cite{havlicek2019supervised, schuld2019quantum}.

Most practical dimensionality reduction pipelines rely on unsupervised methods such as Principal Component Analysis (PCA), which preserve dominant variance directions but do not explicitly preserve class-discriminative structure \cite{jolliffe2016principal}. By contrast, Linear Discriminant Analysis (LDA) constructs supervised projections that maximize between-class separation relative to within-class variation \cite{fisher1936use}. This distinction is especially important in small-data and fine-grained settings, where diagnostically useful variation may not coincide with the directions of the largest global variance \cite{jolliffe2016principal, fisher1936use, barbedo2018impact}. This suggests that representation geometry should be treated not as a preprocessing detail but as a central design variable for small-data QML.

In this work, we study this problem in fine-grained plant phenomics through a hybrid classical-quantum pipeline. Starting from 1280-dimensional deep image embeddings extracted using EfficientNet-B0 \cite{tan2019efficientnet}, we first compress the features into a 64-dimensional PCA space and then apply supervised LDA to obtain an 11-dimensional class-aware latent representation. This \emph{Supervised Latent Restructuring} (SLR) stage is followed by Angle-Aware Latent Rescaling and Quantum Kernel Alignment (QKA), implemented using GPU-accelerated simulation on NVIDIA L40S hardware. Rather than assuming that supervised compression is universally beneficial, we examine how this restructuring changes both latent geometry and downstream behavior across classical and quantum learners.

Our empirical results reveal a nuanced picture. On the one hand, supervised latent restructuring substantially improves the geometric separability of the compressed representation, as measured by both Silhouette analysis and low-dimensional visualization. On the other hand, its downstream effect is strongly model-dependent: some classical learners benefit from the supervised latent space, while stronger nonlinear baselines degrade under the same 11-dimensional bottleneck. QKA in this regime achieves nontrivial multiclass performance but remains challenging under constrained optimization and evaluation budgets. These findings collectively suggest that a qubit-compatible latent space is necessary but not sufficient for strong trainable quantum performance.

The main contribution of this paper is therefore not a claim of quantum superiority, but a careful empirical study of representation geometry under extreme compression in biologically realistic small-data settings. We show that supervised latent restructuring can substantially improve separability while still introducing trade-offs across downstream learners, and we use this setting to examine the practical limits of trainable quantum kernels in a fine-grained 12-class plant pathology task. More broadly, our results argue that future progress in small-data QML will likely require co-design across representation learning, compression, and quantum optimization.

In summary, our contributions are:
\begin{itemize}
    \item We introduce a hybrid pipeline for small-data plant phenomics that combines deep feature extraction, PCA-based compression, supervised latent restructuring via LDA, angle-aware rescaling, and Quantum Kernel Alignment.
    \item We show that supervised latent restructuring substantially improves the geometric separability of the compressed latent space, but introduces a model-dependent trade-off across classical learners.
    \item We evaluate trainable QKA in the resulting 11-dimensional latent space and show that, under constrained optimization budgets, the method achieves nontrivial multiclass performance while remaining substantially below the strongest classical baselines.
    \item We identify representation geometry as a central design variable in small-data quantum learning and highlight the practical difficulty of recovering nonlinear discriminative structure from aggressively compressed biological representations.
\end{itemize}

\section{Related Work}
\label{Sec:rel_work}

\noindent\textbf{Quantum Kernel Alignment and Trainable Quantum Kernels: }
Quantum kernel methods have emerged as a promising direction in quantum machine learning by mapping classical data into high-dimensional Hilbert spaces, where linear separation may become easier. Foundational work by Schuld and Killoran~\cite{schuld2019quantum} and Havl\'{\i}\v{c}ek et al.~\cite{havlicek2019supervised} established the basis for quantum feature maps and kernel-based quantum classification. To address this, recent work has explored trainable quantum kernels and alignment-based objectives that adapt the kernel to the downstream task~\cite{glick2021covariant} rather than relying on fixed encoding alone. However, practical deployment remains challenging because trainable quantum models can suffer from optimization pathologies, including vanishing gradients and poor trainability in larger parameterized circuits~\cite{mcclean2018barren}, motivating the use of gradient-free stochastic methods such as SPSA~\cite{spall1992multivariate}. Our work builds on this line of research by studying Quantum Kernel Alignment (QKA) in a fine-grained multiclass biological setting. In contrast to the predominantly binary or small-class benchmarks often considered in prior work, we focus on a 12-class plant pathology problem under extreme small-data constraints, where both optimization and representation quality become critical.

\noindent\textbf{Latent Space Geometry and Dimensionality Reduction for QML:}
A major bottleneck in quantum machine learning is the encoding of high-dimensional classical data into qubit-limited quantum representations. Because current quantum pipelines can process only a small number of input dimensions, most practical approaches rely on dimensionality reduction before quantum embedding. Principal Component Analysis (PCA) is commonly used for this purpose because it preserves dominant variance directions in a compact form~\cite{jolliffe2016principal}. However, PCA is unsupervised and therefore not explicitly optimized to preserve class-discriminative information. By contrast, Linear Discriminant Analysis (LDA) constructs supervised projections that maximize between-class separation relative to within-class variation~\cite{fisher1936use}. This distinction is especially important in small-data and fine-grained classification settings, where diagnostically meaningful variation may not align with the directions of largest global variance~\cite{jolliffe2016principal, fisher1936use, barbedo2018impact}. Prior work has emphasized that the geometry of the input representation strongly influences downstream kernel performance, including in quantum settings where the induced feature space is highly sensitive to the structure of the encoded data~\cite{schuld2019quantum, havlicek2019supervised}. Motivated by this observation, our work introduces \emph{Supervised Latent Restructuring} (SLR), which combines PCA-based denoising with Linear Discriminant Analysis (LDA) to better preserve class-separating structure before quantum mapping. This positions representation geometry not as a preprocessing detail, but as a central design variable in small-data quantum learning.

\noindent\textbf{Quantum Machine Learning for Agriculture and Plant Phenomics:}
Image-based plant disease recognition has been widely studied using deep learning, especially convolutional neural networks and transfer learning approaches~\cite{mohanty2016using, ferentinos2018deep, mitra2023agrodet}. However, plant disease classification remains sensitive to dataset size, class diversity, and subtle visual overlap between categories~\cite{barbedo2018impact}, making fine-grained phenomics a particularly challenging setting. In such applications, diagnostically meaningful cues may appear as small differences in lesion texture, discoloration, or morphology rather than large global changes~\cite{ferentinos2018deep, barbedo2018impact}. While classical deep learning for plant disease recognition is now well established, quantum machine learning in this domain remains comparatively limited. Our work instead studies a balanced 12-class plant pathology problem in an extremely small-data regime and develops a hybrid quantum kernel pipeline tailored to this setting. Rather than proposing an end-to-end quantum vision model, we focus on how representation compression and latent geometry affect quantum kernel learning in a realistic fine-grained phenomics task.

\noindent\textbf{Gap in Existing Literature:}
Taken together, prior work indicates that the success of quantum kernel methods depends not only on the choice of the quantum feature map or optimization procedure, but also on the geometry of the classical representation being encoded. Yet most existing studies treat dimensionality reduction primarily as a qubit-budget constraint rather than as a discriminative design problem. As a result, the role of supervised latent restructuring before quantum mapping remains underexplored, particularly in fine-grained, small-data biological applications. This limitation is especially relevant in plant phenomics, where diagnostically important differences are often subtle, locally structured, and difficult to preserve under aggressive compression. Our work addresses this gap by showing that explicitly restructuring latent space around class-separating directions substantially improves latent separability and provides a more structured entry point for trainable quantum kernels in a 12-class plant pathology setting.


\section{Method}
\label{Sec:method}

We propose a four-stage pipeline that links high-resolution visual representations in plant phenomics with the low-dimensional structure required for quantum kernel learning, followed by a brief description of the computational implementation. The key idea is to preserve diagnostically meaningful biological variation while transforming deep image features into a compact latent space suitable for quantum encoding. In particular, we address the loss of class-discriminative structure that can arise when high-dimensional embeddings are compressed under extremely small-data conditions.

\paragraph{Deep Feature Extraction:}
We first map each leaf image into a high-dimensional feature space using transfer learning. Specifically, we employ an EfficientNet-B0 backbone~\cite{tan2019efficientnet} pretrained on ImageNet to extract deep embeddings from the Global Average Pooling layer, resulting in a vector $\mathbf{e}_i \in \mathbb{R}^{D}$ for each sample $i$, where $D=1280$. EfficientNet-B0 was selected because it provides compact and expressive pretrained features for fine-grained plant pathology classification. 

In preliminary backbone comparisons conducted over the full dataset, EfficientNet-B0 yielded the highest Silhouette score ($0.031$), compared with $0.0015$ for EfficientNet-B3 and $-0.0021$ for ViT. Although all three scores remain close to zero, EfficientNet-B0 provided the strongest relative class separation and was therefore selected for the downstream compression and quantum-learning pipeline. Although EfficientNet-B0 yielded the highest Silhouette score among the evaluated backbones, the absolute value remained low, indicating that the raw pretrained embedding space was still highly overlapping at the dataset level. This is consistent with the fine-grained nature of the task, which includes subtle symptom differences and compound disease classes. Thus, EfficientNet-B0 was selected not because it produced a cleanly clustered representation, but because it provided the strongest relative starting point for downstream supervised restructuring.

\paragraph{Supervised Latent Restructuring (SLR):}
To make the classical embeddings compatible with the Noisy Intermediate-Scale Quantum  (NISQ)-era quantum encoding, we compress the 1280-dimensional EfficientNet-B0 features into an 11-dimensional latent space through a two-stage transformation.

\begin{enumerate}
    \item \textbf{Unsupervised Denoising:} We first apply Principal Component Analysis (PCA)~\cite{jolliffe2016principal} to the training embeddings and retain the top $d_{\mathrm{pca}}=64$ components. This step reduces dimensionality while filtering redundant variation and high-dimensional noise, thereby stabilizing the subsequent supervised projection:
    \begin{equation}
        \mathbf{z}_{\mathrm{pca},i} = \mathbf{W}_{\mathrm{pca}}^{\top}(\mathbf{e}_i - \boldsymbol{\mu}_{\mathrm{train}})
    \end{equation}
    where $\mathbf{W}_{\mathrm{pca}}$ denotes the PCA projection matrix and $\boldsymbol{\mu}_{\mathrm{train}}$ is the mean embedding computed from the training set.

    \item \textbf{Supervised Discrimination:} We then apply Linear Discriminant Analysis (LDA)~\cite{fisher1936use} in the PCA subspace to obtain a supervised representation that maximizes between-class variance relative to within-class variance~\cite{belhumeur1997eigenfaces}. This yields a final latent representation $\mathbf{z}_i \in \mathbb{R}^{d}$ with $d=11$, corresponding to the maximum discriminative dimensionality available for a $C=12$ class problem, i.e., $d \leq C-1$~\cite{bishop2006pattern}. The resulting latent space is designed to preserve class-separating structure before quantum encoding.
\end{enumerate}

To avoid data leakage, the PCA basis, LDA projection, and all feature scaling parameters are fit only on the training set and then reused for validation and test data without modification. These 11-dimensional latent vectors are subsequently passed to the downstream quantum feature map. Their comparative efficacy against higher-dimensional classical baselines is evaluated in Section~\ref{Sec:experiment}.

\paragraph{Angle-Aware Latent Rescaling (AALR):}
The 11-dimensional supervised latent vectors must be numerically aligned with the periodic structure of rotation-based quantum feature maps. Directly injecting unbounded latent values into phase-encoding circuits can induce unstable behavior, including angular wrapping and reduced kernel sensitivity. Prior work has shown that rotation-based quantum encodings exhibit inherent periodicity and sensitivity to input magnitude \cite{havlicek2019supervised, schuld2019quantum}, motivating controlled rescaling before embedding. To address this, we apply Angle-Aware Latent Rescaling (AALR) (Algorithm~\ref{alg:aalr}), which maps each latent coordinate into a bounded interval $[a,b]$.
Formally, AALR computes per-dimension extrema $z_j^{\min}$ and $z_j^{\max}$ strictly from the training set and applies a component-wise affine transformation:
\begin{equation}
    \tilde{z}_{ij} = a + (b-a)\frac{z_{ij} - z_j^{\min}}{z_j^{\max} - z_j^{\min} + \varepsilon}
\end{equation}
where $\varepsilon = 10^{-8}$ is a stability constant. We evaluated both $[0,1]$ and $[0,\pi]$ as target intervals. Empirically, $[0,1]$ yielded more stable \emph{Simultaneous Perturbation Stochastic Approximation} (SPSA)~\cite{spall1992multivariate} optimization, whereas $[0,\pi]$ was more difficult to train reliably under the available iteration budget. Unless otherwise stated, we report results for the configuration selected based on validation performance.

\begin{algorithm}[t]
\caption{Angle-Aware Latent Rescaling (AALR) with SLR}
\label{alg:aalr}
\KwIn{Training embeddings $\{\mathbf{e}_i\}_{i \in \mathcal{T}}$, evaluation embeddings $\{\mathbf{e}_i\}_{i \in \mathcal{E}}$, intermediate PCA dimension $d_{\mathrm{pca}}$, final LDA dimension $d_{\mathrm{out}}$, target interval $[a,b]$, stability constant $\varepsilon$}
\KwOut{Rescaled supervised latent features $\{\tilde{\mathbf{z}}_i\}$}

Compute PCA on $\{\mathbf{e}_i\}_{i \in \mathcal{T}}$ and project all samples to the intermediate space $\mathbf{z}_{\mathrm{pca},i} \in \mathbb{R}^{d_{\mathrm{pca}}}$\;

Fit LDA on $\{\mathbf{z}_{\mathrm{pca},i}\}_{i \in \mathcal{T}}$ using class labels and project to the final latent space:
\[
\mathbf{z}_i = \mathbf{W}_{\mathrm{lda}}^{\top}(\mathbf{z}_{\mathrm{pca},i} - \boldsymbol{\mu}_{\mathrm{lda}}), 
\qquad \forall i \in \mathcal{T} \cup \mathcal{E}
\]

For each latent dimension $j \in \{1,\dots,d_{\mathrm{out}}\}$, compute training-set extrema:
\[
z_j^{\min} = \min_{i \in \mathcal{T}} z_{ij}, \qquad
z_j^{\max} = \max_{i \in \mathcal{T}} z_{ij}
\]

For each sample $i \in \mathcal{T} \cup \mathcal{E}$ and each dimension $j \in \{1,\dots,d_{\mathrm{out}}\}$, rescale:
\[
\tilde{z}_{ij} = a + (b-a)\frac{z_{ij} - z_j^{\min}}{z_j^{\max} - z_j^{\min} + \varepsilon}
\]

Form the rescaled supervised vector $\tilde{\mathbf{z}}_i = [\tilde{z}_{i1}, \dots, \tilde{z}_{id_{\mathrm{out}}}]^\top$\;

Use $\tilde{\mathbf{z}}_i$ as input to the quantum feature map $\Phi(\tilde{\mathbf{z}}_i)$ for kernel alignment and QSVC classification\;
\end{algorithm}

\paragraph{Quantum Kernel Alignment (QKA):}
The rescaled vectors $\tilde{\mathbf{z}}_i$ serve as inputs to an 11-qubit \texttt{ZZFeatureMap}. To adapt the induced kernel to the biological traits of the 12 disease classes, we perform Quantum Kernel Alignment (QKA) using a \texttt{RealAmplitudes} ansatz~\cite{kandala2017hardware} as a trainable variational layer. For two samples $i$ and $k$, the kernel entry is defined by the transition probability~\cite{havlicek2019supervised}:

\begin{equation}
    K(\tilde{\mathbf{z}}_i, \tilde{\mathbf{z}}_k; \theta) =
\left|
\left\langle \Phi(\tilde{\mathbf{z}}_i; \theta) \mid \Phi(\tilde{\mathbf{z}}_k; \theta) \right\rangle
\right|^2
\end{equation}

where $\Phi(\tilde{\mathbf{z}}, \theta) = U(\theta)V(\tilde{\mathbf{z}})\ket{0}^{\otimes n}$ represents the state preparation incorporating the variational ansatz $U(\theta)$ and the data-encoding circuit $V(\tilde{\mathbf{z}})$. To guide training, we define an ideal class-label similarity matrix whose entries indicate whether two samples belong to the same class:

\begin{equation}
    [K_{\mathrm{target}}]_{ik} =
    \begin{cases}
    1, & y_i = y_k \\
    0, & y_i \neq y_k
    \end{cases}
\label{eq:sim_matrix}
\end{equation}

We then optimize the parameters $\theta$ using SPSA to maximize alignment between the learned kernel and this target similarity structure. This alignment process is intended to adapt the induced quantum feature space to the class structure of the 11-dimensional classical latent manifold.

\section{Experiments}
\label{Sec:experiment}

\subsection{Experimental Setup and Baselines}
\label{Sec:base}
We conduct experiments on the Plant Pathology 2021 fine-grained apple disease classification dataset~\cite{thapa2021plant}. This dataset presents a significant challenge for latent space restructuring due to the presence of ``complex'' and ``multi-disease'' labels, where a single leaf may exhibit overlapping symptoms of multiple pathologies. We treat these as distinct categories within our 12-class framework to test the capacity of the Supervised Latent Restructuring (SLR) pipeline to resolve subtle, co-occurring morphological variations. 

To assess the discriminative effect of the SLR pipeline, we established a benchmarking suite using four classical classifiers: Linear SVM, RBF-kernel SVM, Random Forest, and XGBoost. These models were evaluated under two feature regimes: the 64-dimensional PCA subspace and the 11-dimensional LDA-refined latent space. Following the protocol described in Section~\ref{Sec:method}, all models were trained on a strictly balanced subset of 10 samples per class ($N=120$) and validated on 3 samples per class. For classical baselines, final performance was measured on the full held-out test set of 1,000 samples to assess the generalization of the compressed representations in a few-shot setting.

For QKA, we used the same balanced training and validation subsets, but the final evaluation was performed on a balanced test subset containing up to 30 samples per class, where available. This reduced test protocol was necessary because full kernel evaluation on the complete test set was computationally expensive under the available runtime budget. Validation performance was used to select the downstream QSVC regularization parameter. We report Accuracy, Macro-F1, and Weighted-F1, with Macro-F1 serving as an especially important metric due to class imbalance in the held-out data.

\paragraph{Implementation Details:} All experiments were implemented in Python 3.10 using scikit-learn for PCA, LDA, and classical benchmarking, and PyTorch for deep feature extraction via EfficientNet-B0. Quantum circuits and kernel alignment were developed using Qiskit and Qiskit Machine Learning. Detailed hyperparameter configurations for both the classical baselines and the QKA optimization pipeline are provided in Tables~\ref{tab:qka_hyperparams} and \ref{tab:classical_hyperparams}. Because QKA requires repeated $O(N^2)$ kernel matrix evaluations, training is computationally demanding even in this small-data regime and requires GPU acceleration. We therefore performed simulations on a high-performance computing (HPC) cluster (Atlas on the USDA SCINet) using the \texttt{AerSimulator} with single-precision floating-point arithmetic on NVIDIA L40S GPUs. A full audit of computational resources and GPU-hours is documented in Table~\ref{tab:compute_resources}. This hardware acceleration made iterative optimization over the 12-class pathology task tractable, though kernel evaluation remained the dominant computational bottleneck. 

\begin{table}[htbp]
\centering
\caption{Hyperparameters and Technical Configurations for QKA Optimization.}
\label{tab:qka_hyperparams}
\begin{tabular}{l l}
\toprule
\textbf{Parameter Category} & \textbf{Configuration Details} \\
\midrule
Quantum Circuit & $ZZFeatureMap$ (reps=1), $RealAmplitudes$ (reps=1) \\
Register Size & 11 Qubits (one per LDA-refined feature) \\
Encoding Range & $[0, 1]$ (via MinMaxScaler) \\
\addlinespace
Optimizer (SPSA) & $maxiter=30$, $learning\_rate=0.02$, $perturbation=0.05$ \\
SPSA Stability & $blocking=True$, $allowed\_increase=0.002$, $resamplings=1$ \\
Loss Function & Support Vector Classifier (\texttt{svc\_loss}) \\
\addlinespace
Model Selection & Grid search over $C \in \{0.1, 1.0, 10.0\}$ via validation Macro-F1 \\
Simulation Backend & \texttt{AerSimulator} (Statevector on NVIDIA L40S GPU) \\
Numerical Precision & Single-precision (FP32) \\
Software Stack & Qiskit 1.x, Qiskit Machine Learning, Qiskit Aer \\
\bottomrule
\end{tabular}
\end{table}

\begin{table}[htbp]
\centering
\caption{Hyperparameters for Classical Baseline Models.}
\label{tab:classical_hyperparams}
\begin{tabular}{l l}
\toprule
\textbf{Model} & \textbf{Hyperparameter Configuration} \\
\midrule
Linear SVM & $C=1.0$, Linear Kernel, \texttt{random\_state=42} \\
RBF SVM & $C=1.0$, $\gamma=\text{``scale''}$, RBF Kernel \\
Random Forest & $n\_estimators=300$, Max Depth: None, \texttt{n\_jobs=-1} \\
XGBoost & $n\_estimators=300$, $\eta=0.05$, Max Depth: 6, \texttt{objective=``multi:softmax''} \\
\bottomrule
\end{tabular}
\end{table}

\begin{table}[htbp]
\centering
\caption{Computational Resources and Execution Environment.}
\label{tab:compute_resources}
\begin{tabular}{l l}
\toprule
\textbf{Category} & \textbf{Specification} \\
\midrule
High-Performance Compute & GPU-enabled nodes featuring NVIDIA L40S GPUs \\
Resource Allocation & 64 GB system memory per optimization job \\
Workload Management & SLURM scheduling via persistent \texttt{nohup} sessions \\
Max Job Runtime & 24 hours per independent QKA task \\
\addlinespace
Development + Production & $\sim$209.7 GPU-hours (8 days, 17 hours, 46 minutes) \\
Exploration & $\sim$200 additional GPU-hours \\
\addlinespace
Local Preprocessing & Dell Precision (Intel i9-13950HX, 64 GB RAM) \\
Software Environment & Windows 11 Enterprise (Local) / Linux (Atlas Cluster) \\
Operating System & Rocky Linux 9.1 (on Atlas Cluster) \\
Python & 3.10.20 \\
Qiskit & 1.3.2 / Qiskit Machine Learning 0.8.4 \\
PyTorch & 2.10.0+cu128 \\
scikit-learn & 1.7.2 \\
\bottomrule
\end{tabular}
\end{table}

\subsection{Results}
\label{Sec:res}
We begin our analysis by evaluating the structural properties of the feature space across different representational tiers to validate the geometric effectiveness of the Supervised Latent Restructuring (SLR) pipeline.

\paragraph{Representation Geometry Under Supervised Latent Restructuring:}


To quantify the effect of our Supervised Latent Restructuring (SLR) framework, we computed Silhouette coefficients across three representational tiers. Raw embeddings and unsupervised PCA-64 projections yielded scores near or below zero ($0.003$ and $-0.006$, respectively), indicating substantial class inter-mixing. In contrast, the supervised LDA-11 space achieved a markedly higher score of $0.197$, showing that SLR transforms a highly overlapping feature space into a substantially more structured latent manifold. The transition from a negative to a positive Silhouette coefficient signifies a fundamental shift from a regime where intra-class distance exceeds inter-class distance to one where meaningful class boundaries emerge. This indicates that the 11-dimensional supervised bottleneck is not merely a reduction in volume, but a purposeful distillation of discriminative structure. Although preliminary comparisons suggested that applying LDA directly to the raw embeddings could yield stronger apparent separability, we retained the PCA$\rightarrow$LDA pipeline because it provides a more controlled compression stage prior to supervised restructuring, which is better aligned with the qubit-limited and small-data constraints of the downstream quantum setting.

\begin{figure}[htbp]
     \centering
     \begin{subfigure}[b]{0.3\textwidth}
         \centering
         \includegraphics[width=\textwidth]{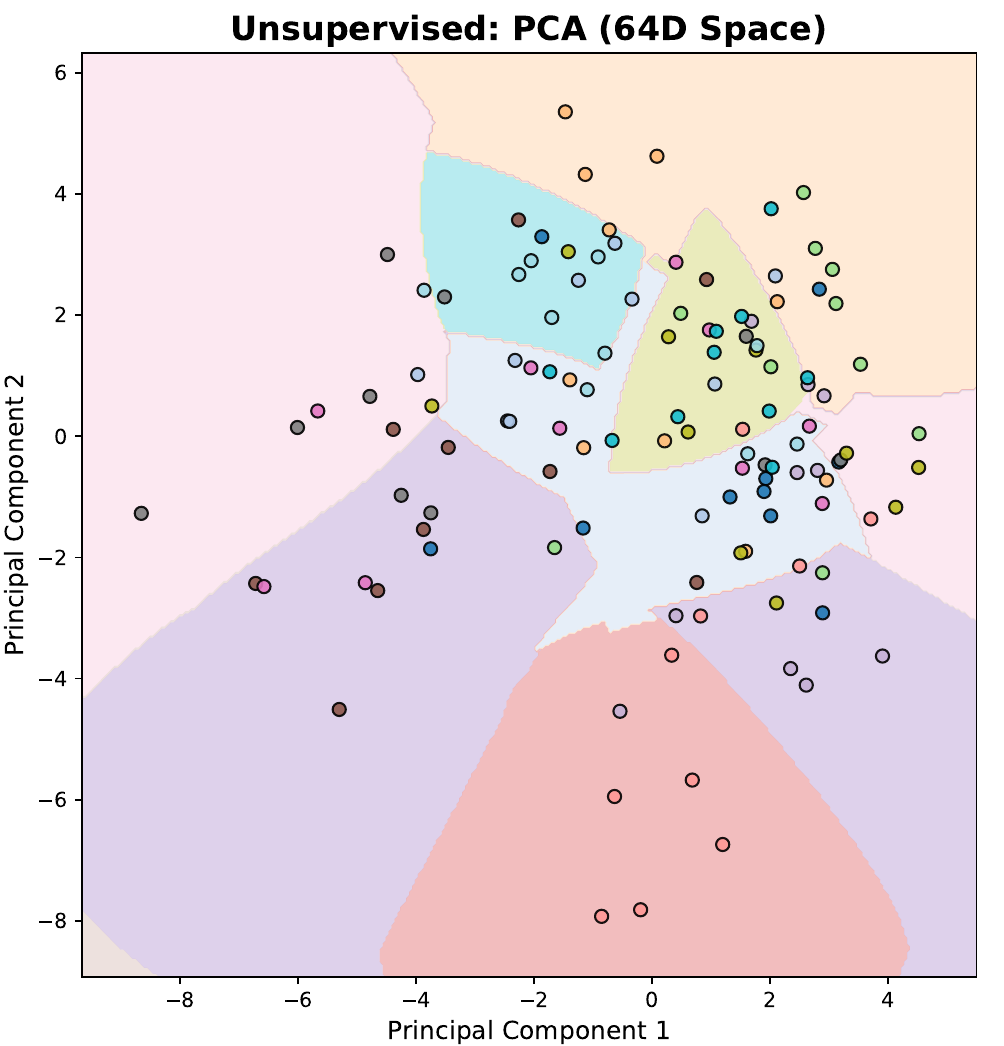} 
         \caption{Unsupervised PCA-64}
         \label{fig:pca_64_geom}
     \end{subfigure}
     \begin{subfigure}[b]{0.36\textwidth}
         \centering
         \includegraphics[width=\textwidth]{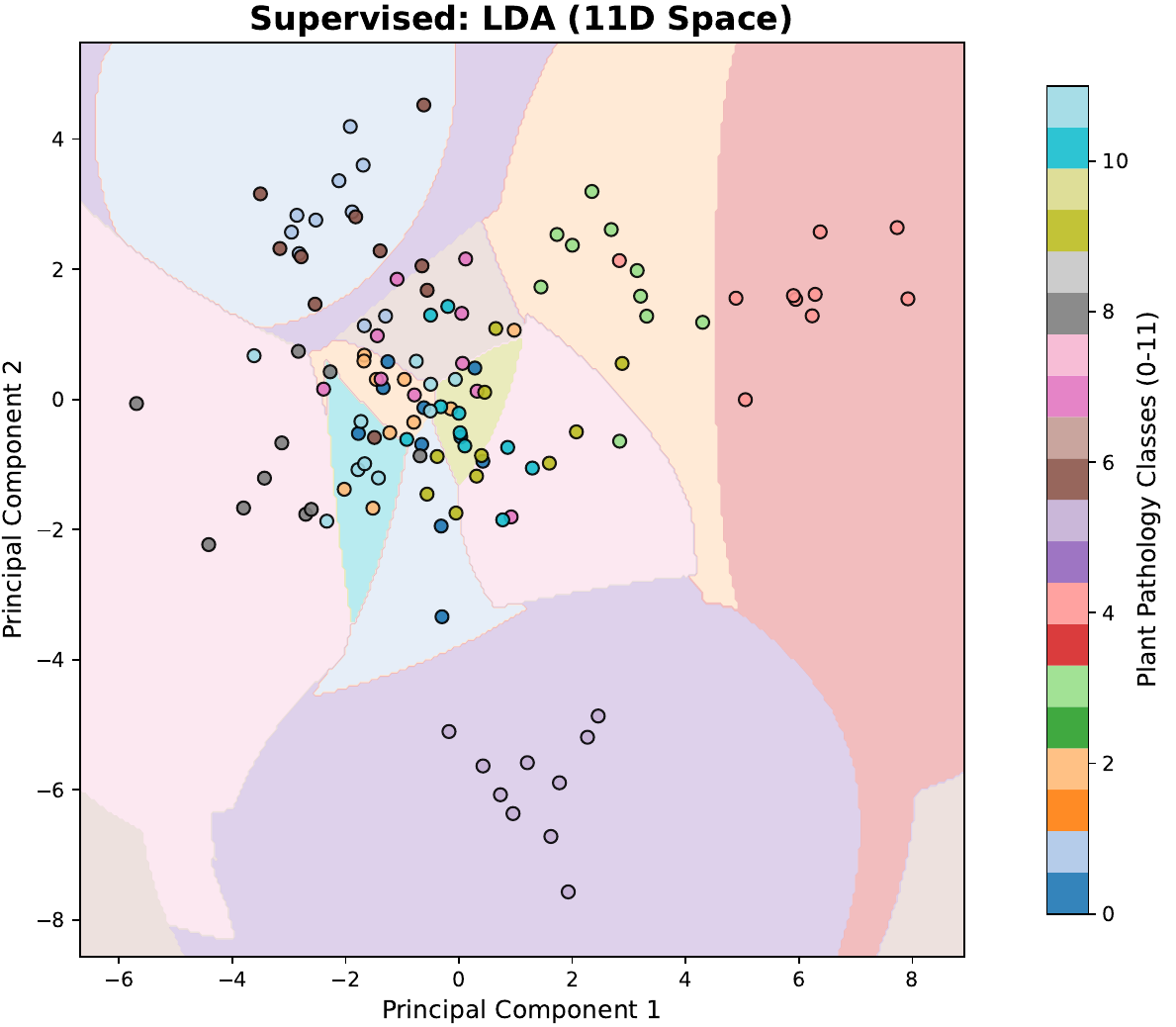}
         \caption{Supervised LDA-11 (SLR)}
         \label{fig:lda_11_geom}
     \end{subfigure}
     \caption{Latent space geometries for the 12-class pathology task. (a) Unsupervised PCA retains variance but exhibits high class overlap. (b) Supervised restructuring (SLR) yields distinct clusters, providing the foundation for QKA.}
     \label{fig:pca_lda}
\end{figure}

We further assessed the qualitative effect of SLR by projecting both PCA-64 and PCA-64 $\rightarrow$ LDA-11 representations into two dimensions for visualization (Fig.~\ref{fig:pca_lda}). While the PCA projection exhibits strong class overlap, the LDA projection reveals the emergence of more compact and separated class groupings. This visual trend is consistent with the Silhouette analysis and indicates that the supervised latent space is better organized around class-discriminative directions.

\begin{figure}[t]
     \centering
     \begin{subfigure}[b]{0.45\textwidth}
         \centering
         \includegraphics[width=\textwidth]{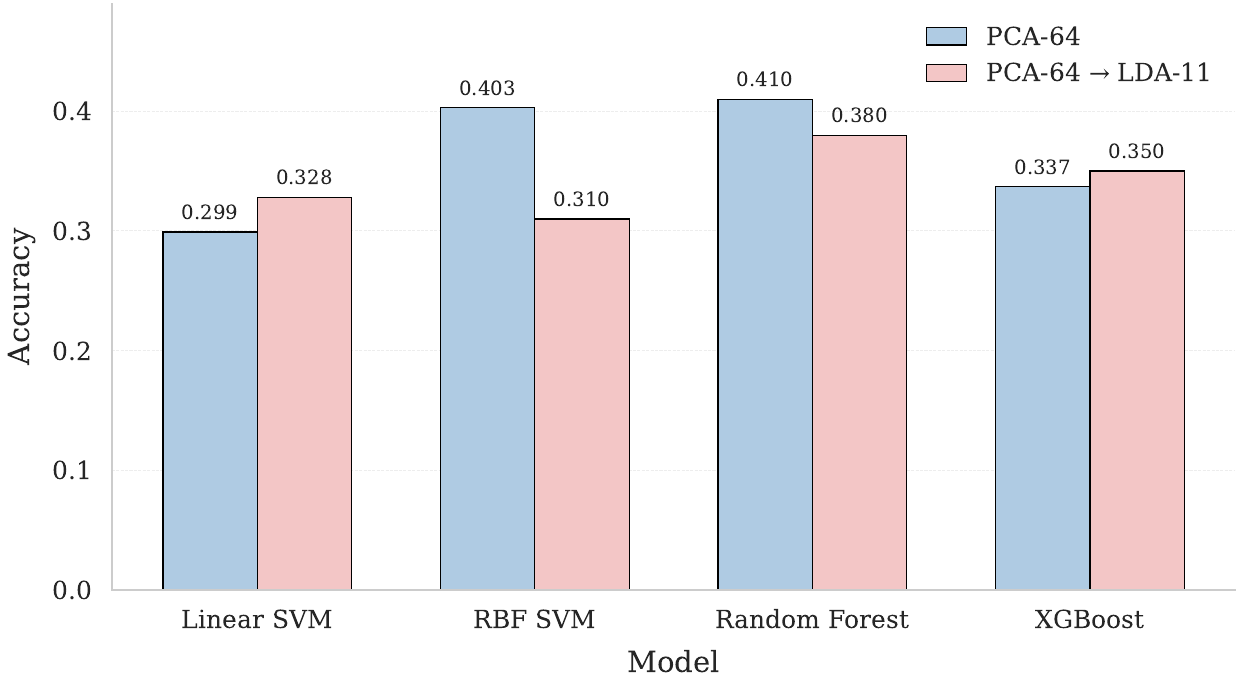}
         \label{fig:pca_64}
     \end{subfigure}
     \hfill
     \begin{subfigure}[b]{0.45\textwidth}
         \centering
         \includegraphics[width=\textwidth]{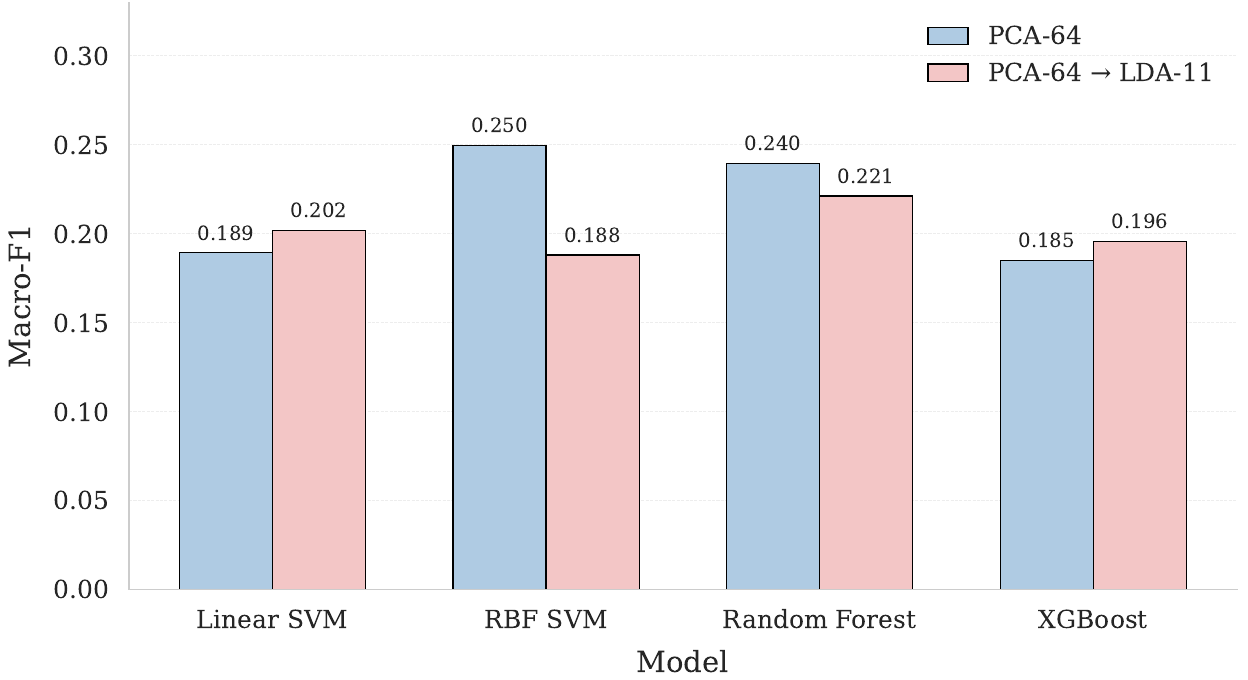}
         \label{fig:lda_11}
     \end{subfigure}
    \caption{Classical Benchmarking under Dimensionality Constraints. (a) Accuracy and (b) Macro-F1 for the 12-class pathology task across two representation tiers. The transition to LDA-11 reveals a divergent trend: linear-biased models improve through class-aware alignment, while flexible non-linear learners lose discriminative ``texture.'' This 11D bottleneck provides a challenging target for QKA.}
     \label{fig:latent_comparison}
\end{figure}

\paragraph{Classical Baselines Under Compression:}
The empirical comparisons in Fig.~\ref{fig:latent_comparison} across the two representation spaces- the original 64-dimensional PCA space and the compressed PCA $\rightarrow$ LDA-11 space- reveal that supervised latent restructuring does not yield uniform gains across all learners; instead, its performance depends heavily on model bias.
Linear SVM and XGBoost improved after LDA restructuring, with Linear SVM accuracy increasing from 0.299 to 0.328 and XGBoost from 0.337 to 0.350. The same pattern appears in Macro-F1, where Linear SVM improved from 0.189 to 0.202 and XGBoost from 0.185 to 0.196. By contrast, RBF-SVM and Random Forest degraded under the compressed 11-dimensional representation. RBF-SVM accuracy dropped from 0.403 to 0.310, with Macro-F1 decreasing from 0.250 to 0.188, while Random Forest accuracy declined from 0.410 to 0.380. 

\begin{figure}[t]
    \centering
    \includegraphics[width=0.65\linewidth]{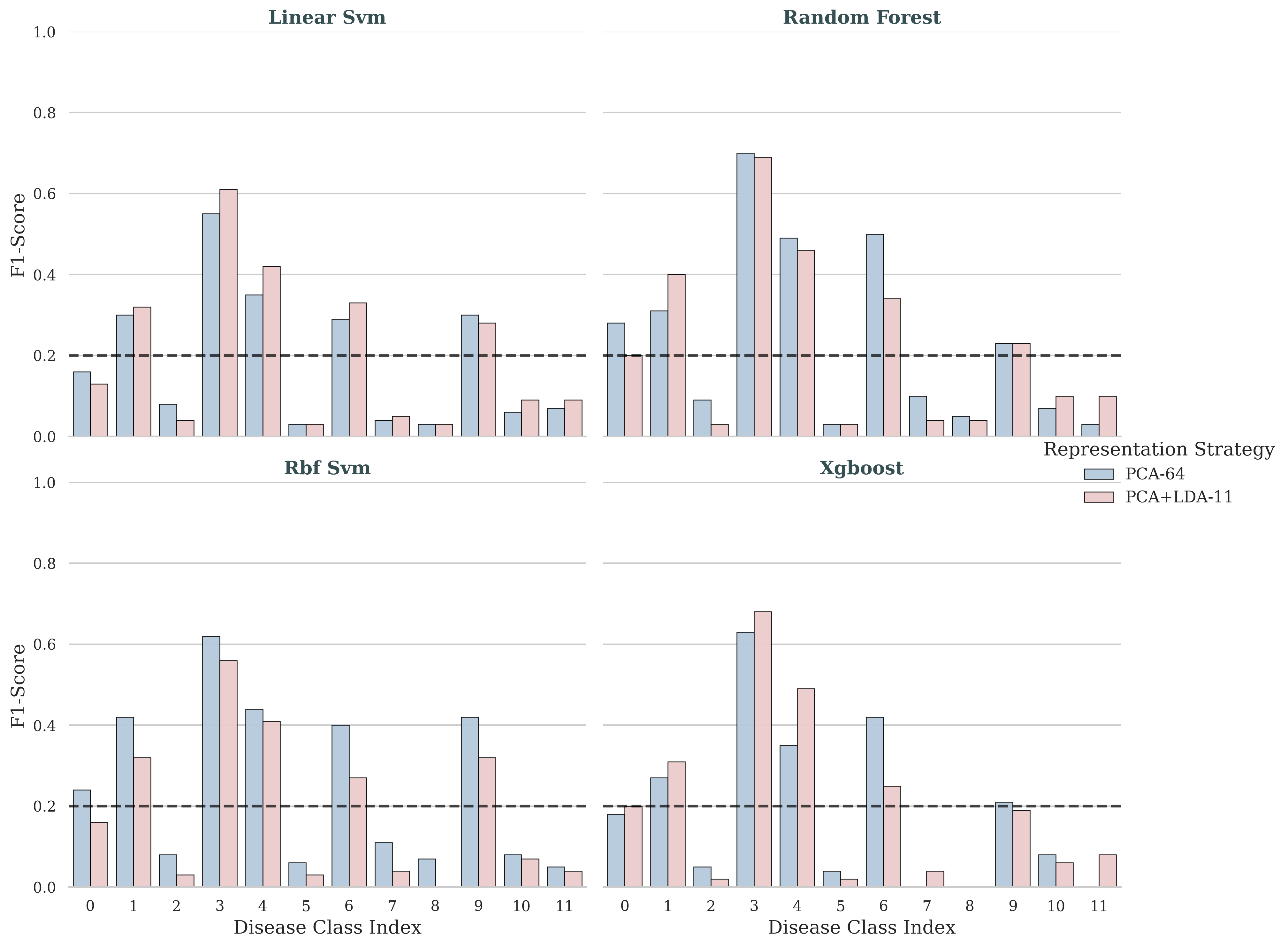}
    \caption{Class-wise $F_1$ comparison across classical baselines under PCA-64 and PCA$\rightarrow$LDA-11 representations. The figure highlights the supervised compression trade-off: the compressed supervised latent space improves some learners, but substantially degrades high-capacity nonlinear models on several minority and visually similar classes.}
    \label{fig:bottleneck}
\end{figure}

The class-wise $F_1$ analysis in Fig.~\ref{fig:bottleneck} makes the supervised compression trade-off more explicit. While the SLR pipeline ($1280 \rightarrow 64 \rightarrow 11$) improves the representation for some learners, it also reduces the discriminative richness available to high-capacity nonlinear models. In the 64-dimensional PCA space, RBF-SVM and Random Forest exploit higher-dimensional variation to separate complex disease boundaries more effectively. After compression to the 11-dimensional LDA space, these models degrade substantially, including near-zero $F_1$ scores for several minority and visually similar classes. This pattern suggests that aggressive compression strengthens compact class-aware structure while also removing fine-grained variation that certain classical nonlinear learners rely on.

These results directly motivate the quantum phase of our study. The classical baselines show that PCA-64 remains a strong benchmark, while the LDA-11 space provides a more structured and qubit-compatible representation that is nevertheless difficult for strong nonlinear classical models. We therefore evaluate whether QKA can exploit this compressed latent geometry more effectively under qubit-limited constraints.

\paragraph{Quantum Kernel Alignment on the Supervised Latent Space:}


When evaluating Quantum Kernel Alignment (QKA) within the compressed 11-dimensional latent geometry described in Table~\ref{tab:qka_smalltest}, the optimized downstream Quantum Support Vector Classifier (QSVC) achieved a validation accuracy of $19.4\%$ and a validation Macro-F1 score of $0.163$ ($C = 1.0$). On the balanced test subset, the corresponding model achieved $23.6\%$ accuracy, $0.212$ Macro-F1, and $0.262$ weighted F1-score.

\begin{table}[ht]
\centering
\caption{QKA results on the supervised PCA$\rightarrow$LDA-11 latent space. Validation selects the QSVC regularization parameter. Test results are reported on a balanced subset containing up to 30 samples per class where available.}
\label{tab:qka_smalltest}
\begin{tabular}{lccccc}
\toprule
\textbf{Setting} & \textbf{Val Acc} & \textbf{Val F1} & \textbf{Test Acc} & \textbf{Test F1} & \textbf{Wt. F1} \\
\midrule
QKA + PCA$\rightarrow$LDA-11 ($C=1$) & 0.194 & 0.163 & 0.236 & 0.212 & 0.262 \\
\bottomrule
\end{tabular}%
\end{table}

While these results trail the classical PCA-64 benchmarks, they demonstrate that the quantum kernel extracts discriminative structure from an extreme few-shot signal (10 samples/class). Given the high complexity of the 12-class fine-grained disease framework and the tight constraints of the optimization budget, this performance indicates a nontrivial mapping into the quantum Hilbert space that partially recovers the class-separating features lost during classical LDA compression.

At the class level, performance remained uneven. The optimized quantum kernel achieved relatively strong signal recovery for Class $3$ (F1 = $0.57$) and moderate performance for Class $1$ (F1 = $0.39$), but several other classes remained difficult, with low or near-zero F1-scores. Ultimately, these results indicate that QKA captures nontrivial multiclass structure in the compressed 11-dimensional latent space, but remains substantially below the strongest classical baselines under the current optimization budget.

\subsection{Discussion}
\label{Sec:diss}
The results show that supervised latent restructuring is best understood as a geometry-shaping operation rather than a uniformly beneficial compression strategy. The PCA $\rightarrow$ LDA transition substantially increases latent-space separability, as seen in both the Silhouette analysis and the two-dimensional projections, confirming that the supervised step reorganizes the compressed representation around class-discriminative directions \cite{fisher1936use}. However, the downstream effects are clearly learner-dependent. Linear SVM and XGBoost benefit from the more structured latent space, whereas RBF-SVM and Random Forest degrade under the same 11-dimensional bottleneck. This indicates that aggressive compression can simplify some decision boundaries while simultaneously removing higher-order variation that stronger nonlinear learners can exploit \cite{cortes1995support, scholkopf2002learning, breiman2001random, chen2016xgboost}.

This trade-off is especially relevant in fine-grained plant phenomics, where diagnostically meaningful differences often appear as subtle changes in lesion texture, discoloration, or morphology rather than large global shifts \cite{mohanty2016using, ferentinos2018deep, barbedo2018impact}. The PCA-64 representation preserves a richer form of variation \cite{jolliffe2016principal}, which appears advantageous for flexible nonlinear classical models. By contrast, the PCA $\rightarrow$ LDA-11 representation strengthens compact class-aware structure, consistent with the supervised objective of LDA \cite{fisher1936use}, but also removes variation that stronger nonlinear models can exploit in the higher-dimensional PCA-64 space. In this sense, the classical baselines do not support a simple claim that supervised compression is always better; instead, they show that the value of compression depends on the inductive bias of the learner operating on the latent space.

The QKA results place an additional constraint on this picture. Although the supervised latent space is geometrically cleaner and qubit-compatible, trainable quantum kernel learning in this regime remains challenging under constrained optimization and compute budgets. As shown in Table~\ref{tab:qka_smalltest}, QKA achieves nontrivial multiclass performance on the balanced test subset, but it remains substantially below the strongest classical baselines. This suggests that improved latent geometry alone is not sufficient to guarantee strong trainable quantum performance. A compressed class-aware representation may be necessary for qubit-limited learning, but the induced quantum kernel can still remain optimization-limited, especially in multiclass small-data settings where the training objective is noisy and kernel evaluations are computationally expensive.


Inspection of the misclassified test images suggests that many failures are visually plausible rather than arbitrary. Errors are concentrated among classes with overlapping lesion morphology and among mixed-disease labels, indicating that the compressed latent space often preserves coarse symptom cues while losing the finer-grained structure needed to distinguish subtle co-occurrence patterns. Additional failures occur in samples with mild or sparse symptoms, partial leaves, cluttered backgrounds, shadows, or strong illumination changes, where the disease signal occupies only a small portion of the image. These patterns are consistent with the quantitative results and suggest that aggressive compression preserves partial disease evidence, but not always the full fine-grained structure required for robust multiclass separation.
In summary, the results support three main observations. 

\begin{itemize}
    \item Supervised latent restructuring substantially improves the geometric separability of the compressed representation. 
    \item This restructuring introduces a model-dependent trade-off in classical learning, benefiting some learners while weakening others. 
    \item QKA on the 11-dimensional latent space remains challenging under constrained optimization and evaluation budgets.
\end{itemize}

 Thus, while SLR produces a more structured and qubit-compatible representation, latent geometry alone is not sufficient to guarantee strong trainable quantum performance in this regime. From this perspective, the present results should not be interpreted as evidence of quantum superiority, but rather as a careful empirical study of where the current bottlenecks lie.

These observations point to several directions for future work. One direction is to study more stable alignment objectives and optimization strategies for trainable quantum kernels under tight evaluation budgets. Another is to compare multiple quantum feature maps and trainable ansatz choices on the same supervised latent space. More broadly, our results suggest that future progress in small-data QML for biological sensing will likely require co-design across representation learning, compression, and quantum optimization, rather than improvements in any one stage alone.

\section{Conclusion}

This work investigated how representation geometry dictates learning behavior under extreme dimensionality compression in fine-grained plant phenomics. We introduced a hybrid framework integrating deep feature extraction, PCA-driven denoising, LDA-based supervised latent restructuring, angle-aware rescaling, and trainable Quantum Kernel Alignment (QKA). Our empirical evaluation reveals that while supervised latent restructuring enhances the geometric separability of the compressed space, it introduces a stark classical trade-off: linear learners benefit from the structured latent topology, whereas high-capacity non-linear baselines lose the high-dimensional variance necessary to resolve fine-grained disease boundaries.

This classical limitation directly establishes the necessity of the quantum phase. By projecting the severely compressed 11-dimensional features into a high-dimensional quantum Hilbert space, QKA demonstrates a nontrivial capacity to extract discriminative structure and recover signal that classical non-linear models completely lose under compression. While constrained optimization budgets currently keep these quantum metrics below the uncompressed classical benchmarks, the results validate the theoretical viability of using quantum kernels to bypass classical geometric bottlenecks in qubit-limited regimes. This identifies representation geometry and quantum optimization as deeply coupled variables that must be advanced jointly to achieve practical quantum advantage in biological sensing.

Shifting toward a deployment paradigm, this methodology serves as a foundational step toward resource-efficient AI for smart agriculture. By demonstrating that meaningful diagnostic signals can be extracted from extreme few-shot regimes (10 samples per class), this framework lays the groundwork for deploying precision diagnostics in regions where agricultural data collection is logistically or financially prohibitive. To ensure its responsible development, we note that this system is intended strictly as a decision-support tool rather than an autonomous actor, minimizing the risk of misdiagnosis. Furthermore, while our current reliance on high-performance classical simulators carries a notable computational footprint, this exploratory phase is essential for benchmarking algorithms before transitioning them to energy-efficient, native NISQ-era quantum hardware.

\section*{Declaration of Generative AI and AI-Assisted Technologies in the Writing Process }

During the preparation of this manuscript, the authors used generative AI tools solely for linguistic refinement, including improvements to grammar, clarity, and prose flow. The authors retained full oversight throughout the process and take complete responsibility for the technical accuracy, mathematical derivations, and empirical integrity of the final work.

\bibliographystyle{unsrt}
\bibliography{references}

@article{havlicek2019supervised,
  title={Supervised learning with quantum-enhanced feature spaces},
  author={Havlíček, Vojtěch and Córcoles, Antonio D. and Temme, Kristan and Harrow, Aram W. and Kandala, Abhinav and Chow, Jerry M. and Gambetta, Jay M.},
  journal={Nature},
  volume={567},
  pages={209--212},
  year={2019}
}

@article{schuld2019quantum,
  title={Quantum machine learning in feature Hilbert spaces},
  author={Schuld, Maria and Killoran, Nathan},
  journal={Physical Review Letters},
  volume={122},
  number={4},
  pages={040504},
  year={2019}
}

@article{glick2021covariant,
    author = "Glick, Jennifer R. and Gujarati, Tanvi P. and Corcoles, Antonio D. and Kim, Youngseok and Kandala, Abhinav and Gambetta, Jay M. and Temme, Kristan",
    title = "{Covariant quantum kernels for data with group structure}",
    eprint = "2105.03406",
    archivePrefix = "arXiv",
    primaryClass = "quant-ph",
    doi = "10.1038/s41567-023-02340-9",
    journal = "Nature Phys.",
    volume = "20",
    number = "3",
    pages = "479--483",
    year = "2024"
}

@article{mcclean2018barren,
  title={Barren plateaus in quantum neural network training landscapes},
  author={McClean, Jarrod R and Boixo, Sergio and Smelyanskiy, Vadim N and Babbush, Ryan and Neven, Hartmut},
  journal={Nature communications},
  volume={9},
  number={1},
  pages={4812},
  year={2018},
  publisher={Nature Publishing Group UK London},
  doi = {https://doi.org/10.1038/s41467-018-07090-4}
}

@article{fisher1936use,
  title={The use of multiple measurements in taxonomic problems},
  author={Fisher, Ronald A.},
  journal={Annals of Eugenics},
  volume={7},
  number={2},
  pages={179--188},
  year={1936}
}

@article{jolliffe2016principal,
  title={Principal component analysis: a review and recent developments},
  author={Jolliffe, Ian T. and Cadima, Jorge},
  journal={Philosophical Transactions of the Royal Society A},
  volume={374},
  number={2065},
  pages={20150202},
  year={2016}
}

@article{cortes1995support,
  title={Support-vector networks},
  author={Cortes, Corinna and Vapnik, Vladimir},
  journal={Machine Learning},
  volume={20},
  number={3},
  pages={273--297},
  year={1995}
}

@article{belhumeur1997eigenfaces,
  title={Eigenfaces vs. fisherfaces: Recognition using class specific linear projection},
  author={Belhumeur, Peter N. and Hespanha, Jo{\~a}o P. and Kriegman, David J.},
  journal={IEEE Transactions on pattern analysis and machine intelligence},
  volume={19},
  number={7},
  pages={711--720},
  year={1997},
  publisher={IEEE}
}

@book{bishop2006pattern,
  author    = {Bishop, Christopher M.},
  title     = {Pattern Recognition and Machine Learning},
  publisher = {Springer},
  year      = {2006}
}

@book{scholkopf2002learning,
  title={Learning with Kernels: Support Vector Machines, Regularization, Optimization, and Beyond},
  author={Sch{\"o}lkopf, Bernhard and Smola, Alexander J.},
  publisher={MIT Press},
  year={2002}  
}

@article{breiman2001random,
  title={Random forests},
  author={Breiman, Leo},
  journal={Machine Learning},
  volume={45},
  number={1},
  pages={5--32},
  year={2001},
  publisher={Springer}
}

@article{mitra2023agrodet,
  title={{aGRodet 2.0: An automated real-time approach for multiclass plant disease detection}},
  author={Mitra, Alakananda and Mohanty, Saraju P and Kougianos, Elias},
  journal={SN Computer Science},
  volume={4},
  number={5},
  pages={657},
  year={2023},
  publisher={Springer}
}

@inproceedings{chen2016xgboost,
  title={XGBoost: A scalable tree boosting system},
  author={Chen, Tianqi and Guestrin, Carlos},
  booktitle={Proceedings of the 22nd ACM SIGKDD International Conference on Knowledge Discovery and Data Mining},
  pages={785--794},
  year={2016}
}

@article{article,
author = {Li, Ruiheng and Wang, Xuaner and Cui, Yuzhuo and Xu, Yifei and Zhou, Yuhao and Tang, Xuechun and Jiang, Chenlu and Song, Yihong and Dong, Hegan and Yan, Shuo},
year = {2025},
month = {02},
pages = {434},
title = {A Semi-Supervised Diffusion-Based Framework for Weed Detection in Precision Agricultural Scenarios Using a Generative Attention Mechanism},
volume = {15},
journal = {Agriculture},
doi = {10.3390/agriculture15040434}
}

@article{mohanty2016using,
  title={Using deep learning for image-based plant disease detection},
  author={Mohanty, Sharada P. and Hughes, David P. and Salath{\'e}, Marcel},
  journal={Frontiers in Plant Science},
  volume={7},
  pages={1419},
  year={2016}
}

@article{ferentinos2018deep,
  title={Deep learning models for plant disease detection and diagnosis},
  author={Ferentinos, Konstantinos P.},
  journal={Computers and Electronics in Agriculture},
  volume={145},
  pages={311--318},
  year={2018}
}

@article{barbedo2018impact,
  title={Impact of dataset size and variety on the effectiveness of deep learning and transfer learning for plant disease classification},
  author={Barbedo, Jayme Garcia Arnal},
  journal={Computers and Electronics in Agriculture},
  volume={153},
  pages={46--53},
  year={2018}
}

@inproceedings{tan2019efficientnet,
  title={Efficientnet: Rethinking model scaling for convolutional neural networks},
  author={Tan, Mingxing and Le, Quoc},
  booktitle={International conference on machine learning},
  pages={6105--6114},
  year={2019},
  organization={PMLR}
}

@article{spall1992multivariate,
  title={Multivariate stochastic approximation using a simultaneous perturbation gradient approximation},
  author={Spall, James C},
  journal={IEEE transactions on automatic control},
  volume={37},
  number={3},
  pages={332--341},
  year={1992},
  publisher={IEEE}
}

@article{kandala2017hardware,
  title={Hardware-efficient variational quantum eigensolver for small molecules and quantum magnets},
  author={Kandala, Abhinav and Mezzacapo, Antonio and Temme, Kristan and Mauser, Kristine and Brink, Markus and Chow, Jerry M and Gambetta, Jay M},
  journal={Nature},
  volume={549},
  number={7671},
  pages={242--246},
  year={2017},
  publisher={Nature Publishing Group}
}

@inproceedings{thapa2021plant,
  title={The Plant Pathology Challenge 2021: Fine-Grained Classification of Plant Diseases},
  author={Thapa, Rishi and Zhang, Kai and Snavely, Noah and Belongie, Serge and Khan, Abeer},
  booktitle={Proceedings of the IEEE/CVF Conference on Computer Vision and Pattern Recognition (CVPR) Workshops},
  pages={2777--2781},
  year={2021}
}

\end{document}